# Think First, Assign Next (ThiFAN-VQA): A Two-stage Chain-of-Thought Framework for Post-Disaster Damage Assessment

Ehsan Karimi, Nhut Le, Maryam Rahnemoonfar*

*Abstract*—Timely and accurate assessment of damages following natural disasters is essential for effective emergency response and recovery. Recent AI-based frameworks have been developed to analyze large volumes of aerial imagery collected by Unmanned Aerial Vehicles (UAVs), providing actionable insights rapidly. However, creating and annotating data for training these models is costly and time-consuming, resulting in datasets that are limited in size and diversity. Furthermore, most existing approaches rely on traditional classification-based frameworks with fixed answer spaces, restricting their ability to provide new information without additional data collection or model retraining. Using pre-trained generative models built on in-context learning (ICL) allows for flexible and open-ended answer spaces. However, these models often generate hallucinated outputs or produce generic responses that lack domain-specific relevance. To address these limitations, we propose Think First, Assign Next (ThiFAN-VQA), a two-stage reasoning-based framework for Visual Question Answering (VQA) in disaster scenarios. ThiFAN-VQA first generates structured reasoning traces using chain-of-thought (CoT) prompting and ICL to enable interpretable reasoning under limited supervision. A subsequent answer selection module evaluates the generated responses and assigns the most coherent and contextually accurate answer, effectively improve the model performance. By integrating a custom information retrieval system, domain-specific prompting, and reasoning-guided answer selection, ThiFAN-VQA bridges the gap between zero-shot and supervised methods, combining flexibility with consistency. Experiments on FloodNet and RescueNet-VQA, UAV-based datasets from flood- and hurricane-affected regions, demonstrate that ThiFAN-VQA achieves superior accuracy, interpretability, and adaptability for real-world post-disaster damage assessment tasks.

*Index Terms*—Multi-modal large language models, Visual question answering, Natural disaster assessment, Hallucination, Chain-of-thought, and In-context learning.

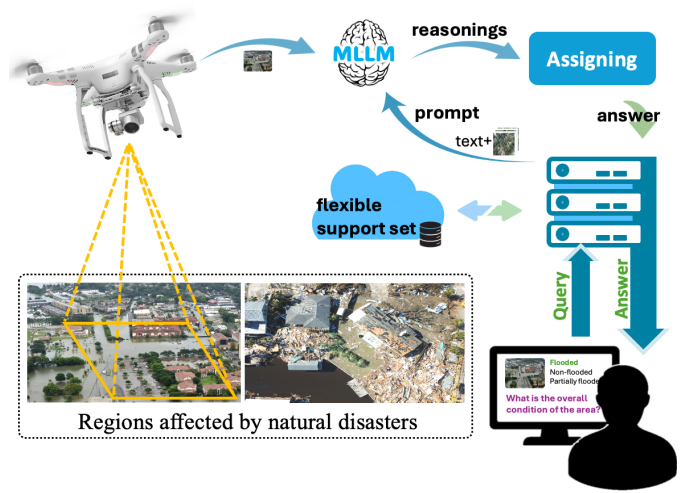

Fig. 1. Overview of the proposed Think First, Assign Next (ThiFAN-VQA) framework, where rescuers query a drone-integrated VQA system to analyze UAV imagery of disaster-affected regions. The MLLM performs reasoning via chain-of-thought prompting and in-context learning, followed by an answer assignment module that ensures accurate and interpretable damage assessment.

## I. INTRODUCTION

IN the immediate aftermath of natural disasters, first responders rely heavily on up-to-date information to assess damage, identify hazards, allocate resources, and reach survivors as quickly as possible. Timely and accurate situational awareness is critical to rescuing operations; however, it has always posed a significant challenge due to the lack of data from the affected area for accurate data-driven planning. Research shows that delays in one phase of disaster response can cascade into longer recovery times by an order of magnitude, highlighting the need for fast, interactive, and data-driven systems to have comprehensive reports from these regions [1], [2]. However, this process is inherently challenging, as access to disaster-stricken areas—particularly after hurricanes or floods—is often severely limited [3], [4]. Consequently, remote sensing technologies have emerged as indispensable tools in modern disaster management, enabling rescue teams to acquire visual information from a wide range of affected areas. Among these technologies, Unmanned Aerial Vehicles (UAVs) play a crucial role in post-disaster assessment by providing high-resolution imagery and offering greater flexibility and responsiveness compared to traditional satellite-based platforms [4], [5]. Nevertheless, in the aftermath of a disaster, UAV missions generate a vast amount of newly acquired visual data from affected regions, and transforming the raw images into meaningful and actionable insights remains a significant challenge [6] and requires to be addressed using automated systems.

Recently, Visual Question Answering (VQA) task has

* Corresponding author (e-mail: maryam@lehigh.edu)

E. Karimi and N. Le are with the Department of Computer Science and Engineering, Lehigh University, Bethlehem, Pennsylvania, 18015, USA (e-mail: ehk224@lehigh.edu and nhl224@lehigh.edu).

M. Rahnemoonfar is with the Department of Computer Science and Engineering, and the Department of Civil and Environmental Engineering, Lehigh University, Bethlehem, Pennsylvania, 18015, USA (e-mail: maryam@lehigh.edu).

emerged that enables the extraction of specific, human-interpretable information directly from disaster imagery [7], [8]. VQA systems allow automated implicit reasoning over visual content in response to natural language questions, enabling efficient interpretation of complex scenes. This task, enables the agent to interact with the system for extracting critical information from the affected regions in an interactive manner (Fig. 1). Despite recent advances in general-purpose VQA models driven by the emergence of multi-modal large language models (MLLMs), applying VQA in the context of natural disaster assessment remains a difficult task [9]. Supervised VQA approaches [9] have shown promise in recent studies. However, their practical adoption in real-world disaster response remains limited due to several persistent challenges. Supervised models, constrained by the scarcity of large-scale and diverse post-disaster datasets, are typically trained on specific disaster events and often fail to generalize to unseen scenarios [9]–[13]. Moreover, some approaches, including Sam-VQA [9], require additional side information, substantially increasing dataset preparation time that is not applicable in urgent situations. To address the scarcity of data and the need for auxiliary information, LLM-/MLLM-based zero-shot frameworks [11], [14], [15] have been proposed by leveraging pretrained models to develop training-free and rapidly deployable solutions. However, their effectiveness can be further enhanced by mitigating the substantial domain gap between the knowledge encoded in models and disaster imagery. Few-shot in-context learning (ICL) has emerged as an effective strategy for narrowing the gap between supervised and zero-shot paradigms [16]. By providing a pretrained model with a set of relevant, retrieved exemplars prior to the main query, ICL can help adapt model behavior to the target domain. Despite its promise, this strategy has not yet been fully explored or optimized for post-disaster scene understanding.

Another major limitation affecting the generalization ability of existing approaches is that most VQA frameworks formulate the task as a classification problem, where the model is constrained to choose an answer only from a fixed set of candidates encountered during training. This closed answer space inherently restricts the model's adaptability and prevents it from handling unseen responses [7], [11], [17]–[26]. To overcome this restriction, text-generative models have recently been explored, allowing the system to generate or select answers from a dynamic and open candidate space, thereby enabling responses that were not explicitly observed during training [14]. While these generative approaches alleviate the limitations of closed-set answer spaces, they often fail to fully leverage the reasoning capabilities of Large Language Models (LLMs). LLMs demonstrate powerful reasoning skills, allowing them to extract salient information from complex inputs through systematic analysis and query interpretation. Moreover, they can explicitly express intermediate reasoning steps prior to producing the final answer that not only improve the output accuracy, but also increase the interpretability of the model. However, these intermediate reasoning steps can sometimes be inconsistent with the final answer, as the output may be influenced by hallucinations or spurious reasoning in MLLMs [1], [27].

To harness the full potential of LLM reasoning and address the reasoning–answer misalignment problem, we propose ThiFAN-VQA – a two-stage reasoning-based answer selection framework for VQA, specifically designed for post-disaster scene understanding in flood- and hurricane-affected regions. Our framework leverages both ICL—which enables the model to infer patterns from provided examples—and Chain-of-Thought (CoT) prompting, which encourages the model to generate step-by-step reasoning before reaching a conclusion. To mitigate the hallucination issue, we introduce a reasoning-based answer selection mechanism that operates after the CoT reasoning generation stage. This mechanism explicitly evaluates and selects the most coherent answer based on the reasoning trace, independently.

As illustrated in Fig. 2, the proposed framework integrates the reasoning capabilities of MLLMs with a domain-adapted information retrieval module, tailored prompting strategies, and a two-stage reasoning–answer selection pipeline. This design allows our approach to achieve performance comparable to supervised methods while maintaining the flexibility, generalization, and zero-shot adaptability characteristic of LLMs. We validate our framework on two UAV-based post-disaster datasets—FloodNet [4] and RescueNet-VQA [8]—capturing flood- and hurricane-affected regions, respectively. Experimental results demonstrate that our method delivers accurate and actionable insights from aerial imagery for real-world disaster assessment tasks. The main contributions of this work are summarized as follows:

- We propose a reasoning-based answer selection framework that combines CoT reasoning with a constrained answer selection stage to enhance interpretability and mitigate hallucination in post-disaster VQA tasks.
- We jointly leverage ICL and CoT techniques for post-disaster scene understanding, enabling robust reasoning and answer generation without the need for fine-tuning or extensive supervision, while dynamically expanding the support set through adaptive example selection.
- We design a question-type-aware information retrieval module that selects semantically relevant image–question–answer exemplars as in-context demonstrations, improving reasoning relevance and performance in flood and hurricane post-disaster assessments.

The remainder of this paper is organized as follows. Section II reviews the related research on post-disaster assessment and VQA. Section III describes the datasets utilized in this study. Section IV presents the proposed framework in detail, while Section V discusses the experimental results and ablation studies. Finally, Section VI concludes the paper and outlines potential directions for future work.

## II. RELATED WORKS

This section reviews the existing literature relevant to our study, encompassing research in remote sensing, post-disaster damage assessment, and VQA. We first provide an overview of VQA, post-disaster assessment, and the challenges of applying VQA to disaster-related scenarios, where models must reason jointly over complex visual and textual information. We then narrow our discussion to two major paradigms in the evolution of VQA frameworks for remote sensing: (1) conventional classification-based approaches, which rely on fixed candidate answer sets, and (2) LLM- and MLLM-based frameworks, which exploit large-scale pretrained models and advanced reasoning mechanisms such as ICL and CoT. Finally, we highlight existing limitations in applying these methods to post-disaster scenarios—such as information loss, lack of domain-specific adaptation, and reasoning–answer misalignment—that motivate the development of our proposed framework.

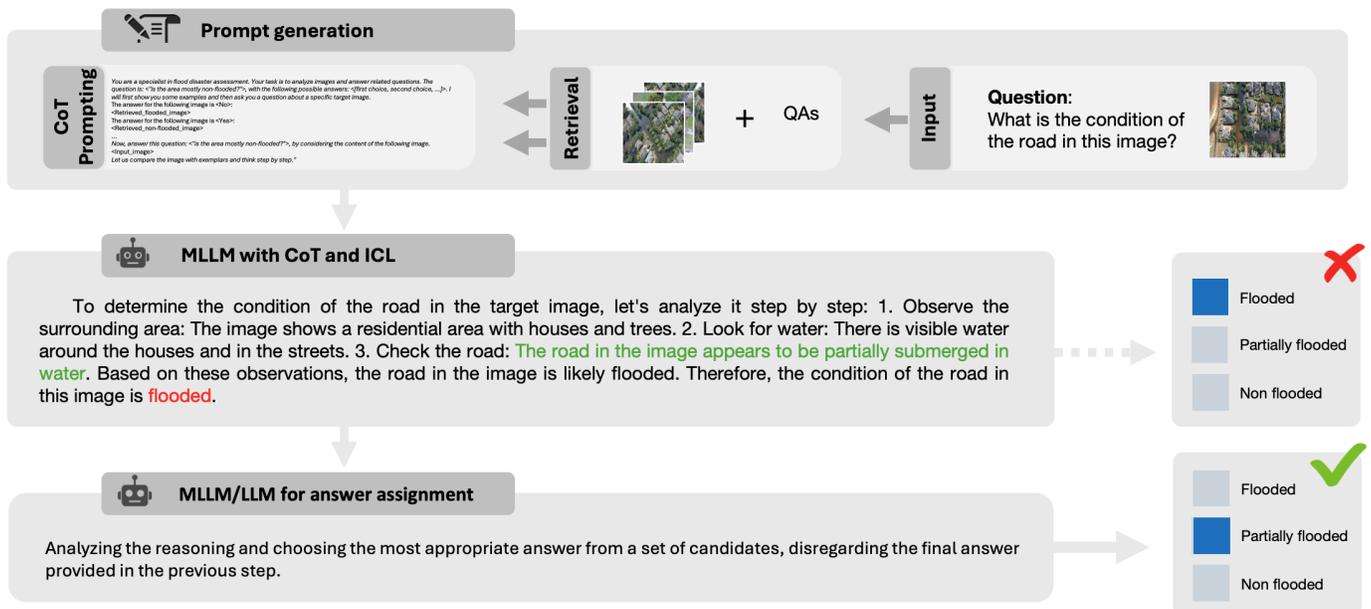

Fig. 2. Overview of the proposed reasoning-based post-disaster VQA framework (ThiFAN-VQA). ThiFAN-VQA integrates MLLMs with a question-type-aware information retrieval module, tailored CoT prompting strategy, and a two-stage reasoning–answer selection pipeline. The reasoning answer assignment stage uses only the reasoning of the CoT to select the final answer independently and reduce the hallucination as well. This design enables an interpretable and few-shot reasoning for UAV-based post-disaster assessment tasks on flood- and hurricane-affected regions.

### A. Post-disaster scene understanding

Recent research in post-disaster analysis has extensively leveraged computer vision methods to localize and quantify structural damage from remote sensing imagery. Early efforts applied classification or detection models to identify damaged regions or count affected buildings [4], [8], [28]. More recent studies advanced toward semantic segmentation [4], [29]–[33] and change detection frameworks that jointly analyze pre- and post-disaster images to capture fine-grained spatial variations [9], [28]–[30], [34], [35]. Beyond pixel-level perception, scene understanding approaches and captioning models [35], [36] have been proposed to produce higher-level, human-interpretable descriptions of areas. However, these models typically rely on task-specific architectures. To overcome this limitation, recent works explore vision-language models (VLMs) and VQA frameworks to unify perception and reasoning, enabling more flexible and query-driven post-disaster assessment.

VQA is a particularly challenging task that requires models to perform joint reasoning over both visual and textual modalities in order to answer questions grounded in image content [14], [16], [37], [38]. While recent advancements in MLLMs have led to substantial progress in general-domain VQA, their application in the context of natural disasters remains an open and challenging problem [15]. This is largely due to the scarcity of labeled data, significant variation in disaster scenarios, and the inherent diversity in task formulations and data distributions [14]. In post-disaster assessment and remote sensing contexts, VQA can be framed either as a classification task–selecting an answer from a predefined set– or as an open-ended generation task addressed by LLMs or MLLMs.

### B. Conventional classification-based VQA frameworks

Early VQA research in remote sensing and post-disaster assessment primarily relied on convolutional neural networks (CNNs) to extract high-level features from aerial or satellite imagery, while recurrent models such as LSTM and GRU were employed to encode textual information. The modality-specific features were subsequently fused through a joint embedding or fusion layer, followed by a classifier to predict the final answer [9], [13], [39]. Later, conventional neural architectures were replaced with Transformer-based models—such as BERT for language and Vision Transformer (ViT) for vision [40]—to capture richer and more expressive representations. In parallel, simple fusion strategies (e.g., dot product) were supplanted by more sophisticated architectures, including multilayer perceptrons and cross-modal attention networks [21]–[23], [41]. These architectural improvements enabled finer-grained interactions between visual and textual modalities, resulting in notable performance gains across VQA benchmarks [37],

[42]. For example, studies such as [26], [43] demonstrated that jointly integrating vision and language features through Transformer-based fusion provides a more effective and unified framework. However, despite these advancements, these classification-based methods still suffer from a closed answer space and limited generalization to unseen datasets or disaster types [10].

*C. LLM-based VQA frameworks*

One of the primary challenges in post-disaster assessment and remote sensing is the scarcity of large-scale annotated datasets, which limits the ability of traditional deep learning models to generalize across diverse disaster scenarios. Leveraging the pretrained knowledge of LLMs offers an effective way to mitigate this limitation by transferring linguistic and reasoning knowledge from large-scale corpora. Moreover, the reasoning capabilities of LLMs have motivated researchers to explore their potential in the domain of natural disaster assessment. For instance, in [11], the authors employ an image captioning model to first generate a textual description of an input image, which is then processed by an LLM to answer the question using the CoT reasoning technique [44]. This zero-shot approach enables strong generalization and rapid deployment in practice. However, as it relies solely on textual captions to convey visual information, it introduces a significant information bottleneck, leading to notable performance degradation. For example, counting questions—already challenging even for supervised models—are particularly problematic, as conventional captioning models typically omit quantitative details such as the number of buildings or damaged structures. This limitation extends to other reasoning types as well, where important spatial or contextual cues are lost in the captioning stage.

The success of vision-language models (VLMs) such as CLIP [45] and BLIP [46], which align fine-grained textual and visual representations, combined with the reasoning ability of LLMs, has led to the development of MLLMs. Models such as Qwen-VL [47], [48], Flamingo [49], and LLaVA [50] have demonstrated remarkable performance across a wide range of multimodal reasoning tasks. Recent studies have increasingly adopted these models as foundational backbones, leveraging their pretrained multimodal knowledge to improve performance in low-resource and domain-specific settings, including natural disaster assessment [14], [51], [52].

Zeshot-VQA [14] addresses the issue of information loss identified in [11] by employing an integrated MLLM in which visual and textual features interact directly within the model. In addition, ZeShot-VQA introduces a selection stage to bridge the gap between the model's general pretraining knowledge and domain-specific expert knowledge. This stage also aligns the model's predicted answers with a set of candidate answers. However, the model still lacks background knowledge specific to the natural disaster domain and dataset, highlighting the need to provide such contextual information to improve performance. However, despite these improvements, the model still lacks contextual knowledge specific to the natural disaster domain and dataset, which limits its adaptability and reasoning accuracy. This contextual information can be effectively incorporated through ICL strategies.

ICL is a powerful technique that leverages the reasoning capabilities of LLMs to enhance performance without requiring parameter updates [53]. Often regarded as a form of few-shot learning [16], [54], ICL operates by presenting the model with a series of exemplars—input–output pairs that implicitly define a selection function—followed by a query input for which the model must generate an output. These exemplars serve as contextual demonstrations, enabling the model to infer patterns and adapt to new tasks through example-based reasoning. Exemplars can be either manually predefined or dynamically selected via retrieval algorithms [53]. Prior studies [55], [56] have demonstrated that the performance of ICL-based systems is highly sensitive to both the selection strategy and the ordering of exemplars, emphasizing the importance of effective context construction in optimizing model reasoning and generalization.

GEOChat [52], TEOChat [51], and LIZAt [6]—the latter trained specifically for reasoning segmentation—address the aforementioned challenges by aggregating multiple Earth observation datasets that encompass a diverse range of question types, including change question answering (QA), temporal QA, and referring expression grounding. These datasets are then used to fine-tune multi-MLLMs, establishing foundation models for aerial imagery understanding. Despite these advancements, such models still exhibit inconsistencies between their generated reasoning and final answers when prompted with CoT reasoning techniques, underscoring a persistent reasoning–answer misalignment issue in MLLM-based reasoning.

To enable fast deployment, adaptability to dynamic or limited datasets, and mitigation of hallucination in disaster scenarios, we propose an ICL-based framework with a reasoning-based answer selection module, leveraging a pretrained MLLM as the backbone. During inference, our approach integrates three key components: (1) a retrieval mechanism that selects contextually relevant exemplars tailored for VQA tasks in natural disaster settings, (2) a cross-modal MLLM that jointly reasons over visual and textual features, and (3) a reasoning-based answer selection module that aligns the model's generated reasoning with candidate answers, thereby reducing hallucination and producing more reliable outputs for post-disaster assessments.

## III. DATA

To train and evaluate our algorithm, we utilize the Flood-Net [4] and the RescueNet-VQA [8] datasets.

FloodNet comprises 2,188 images collected by UAVs during Hurricane Harvey (2017) over affected areas in Fort Bend County, Texas. Fig. 3 shows the sample for VQA in the FloodNet [4]. The dataset includes both flooded and non-flooded regions and contains 7,355 question-answer pairs based on 31 unique queries. These queries are organized into four categories (seven sub-categories):

- Counting: This category includes two types of counting questions. *Simple counting* queries focus on quantities without additional conditions (e.g., *"What is the total*

*number of buildings in the affected area?"*). In contrast, *complex counting* introduces conditional elements, such as *"How many damaged buildings are in this image?"*. Both types are treated as regression tasks.

- Condition Recognition: These questions assess the state of objects within an image, primarily determining flooding conditions. They are divided into yes/no and multiple-choice formats. For example, a yes/no question might be *"Are there any flooded buildings?"* with answers yes, no, while a multiple-choice question could be *"What is the condition of most buildings in this image?"* with possible responses partially flooded, non-flooded, flooded.
- Density Estimation: This category evaluates the density of objects in a given area, with answers classified as low, moderate, high.
- Risk Assessment: These questions assess whether a given area requires urgent intervention, with binary responses yes, no.

The RescueNet-VQA dataset comprises 4,375 UAV captured high-resolution images and 103,192 image-question-answer triplets, captured following Hurricane Michael in Mexico Beach and surrounding areas. It supports VQA tasks including counting, condition recognition, risk analysis, density estimation, level of damage, positional queries, and change detection. The first four categories are structured similarly to their counterparts in the FloodNet but adapted for hurricane-specific scenarios. The remaining categories are as follows:

- Positional queries: Requires high-level scene understanding to reason about the spatial relationships of objects. Example questions include: *"How much damage does the largest building in this image have?"* and *"What is the damage status of the smallest building in this image?"*.
- Change Detection: Focuses on determining whether the number of buildings has changed due to destruction caused by the disaster. These questions are answered solely based on post-hurricane images, as temporal (pre-disaster) images are not available. Example question: *"Is there any change in the number of buildings after the disaster on the scene?"*.
- Level of Damage: Focuses on assessing the overall damage in a scene. Example question: *"How badly damaged is the scene?"*.

## IV. METHODOLOGY

To address the challenges outlined in Section I—including fixed answer spaces, limited dataset availability, and hallucination in LLM-based models—we propose ThiFAN-VQA, a framework designed for rapid deployment, flexible answer generation, and enhanced accuracy. As shown in Fig. 4, the framework integrates ICL and CoT prompting to improve reasoning capability and answer reliability. In addition, it introduces a reasoning-based answer selection module that aligns the model's reasoning process with the final prediction, mitigating hallucination. In this section, we describe the proposed method and detail each component of the ThiFAN-VQA framework.

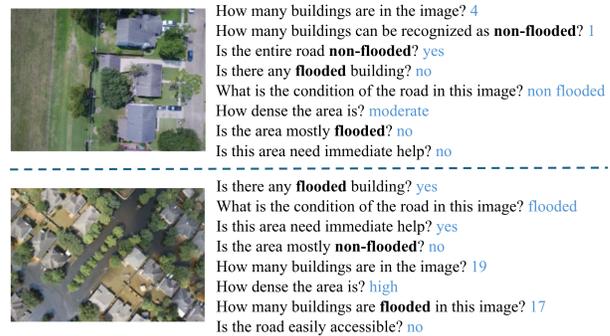

Fig. 3. Image-question-answer triplet samples in FloodNet.

### A. Framework architecture

As shown in Fig. 4, the proposed framework includes four main stages: 1) Support set encoding: extracting embedding vectors of training images and saving them in a vector-based database; 2) Exemplar retrieval: retrieving the demonstration examples for the given paired image-question; 3) Prompt generation and reasoning generation, and 4) Answer selection to map the model reasoning output to one of the candidate answers.

*1) Support set encoding (initial stage):* To facilitate efficient and semantically meaningful retrieval of relevant exemplars during inference, we begin by extracting dense embedding vectors from all training images using the vision encoder of the pre-trained CLIP model [45]. Specifically, we use the ViT-Base variant [40], which consists of 12 transformer layers with a hidden dimension of 768, a projection dimension of 512, and a patch size of 32. This encoder has been shown to capture rich semantic representations that are well-aligned with natural language, making it particularly suitable for vision-language tasks like VQA.

Before feature extraction, all images are resized and normalized according to CLIP's input specifications to ensure consistency with its training distribution. Each resulting image embedding is then stored in a vector-based database, indexed by its corresponding image_id and question_id. This forms structured triplets of (image_embedding, question_id, image_id), which serve as the foundation for the exemplar retrieval process during inference.

By pre-encoding the training set into a searchable vector space, we enable rapid similarity-based retrieval without the need for repeated forward passes through the encoder. This strategy not only improves computational efficiency but also supports flexible and dynamic in-context example selection, which is essential to our ICL-based framework.

*2) Exemplar retrieval:* The proposed framework eliminates the need for a dedicated training phase by utilizing exemplar demonstrations during inference to generate reasoning for a given image–question pair. The process begins with a candidate generation step, where the vector database is filtered based on the input question to retain only samples corresponding to the same query type. For multiple-choice questions, the filtered candidates are further divided into subsets, where each subset contains exemplars with the same answer. In contrast, for open-ended questions, the filtered list is directly passed to the ranking and selection block. In this stage, the input image is encoded using the same vision encoder employed in the earlier phase to extract its embedding vector. Candidate exemplars are then ranked according to their visual similarity to the query image, measured via cosine similarity, computed as:

$$\text{Sim}(I_{\text{input}}, I^{\text{train}}_i) = \frac{g(I\text{input}) \cdot g(I^{\text{support}}_i)}{\|g(I\text{input})\| \cdot \|g(I^{\text{support}}_i)\|} \quad (1)$$

where $g(.)$ denotes the embedding function provided by the vision encoder.

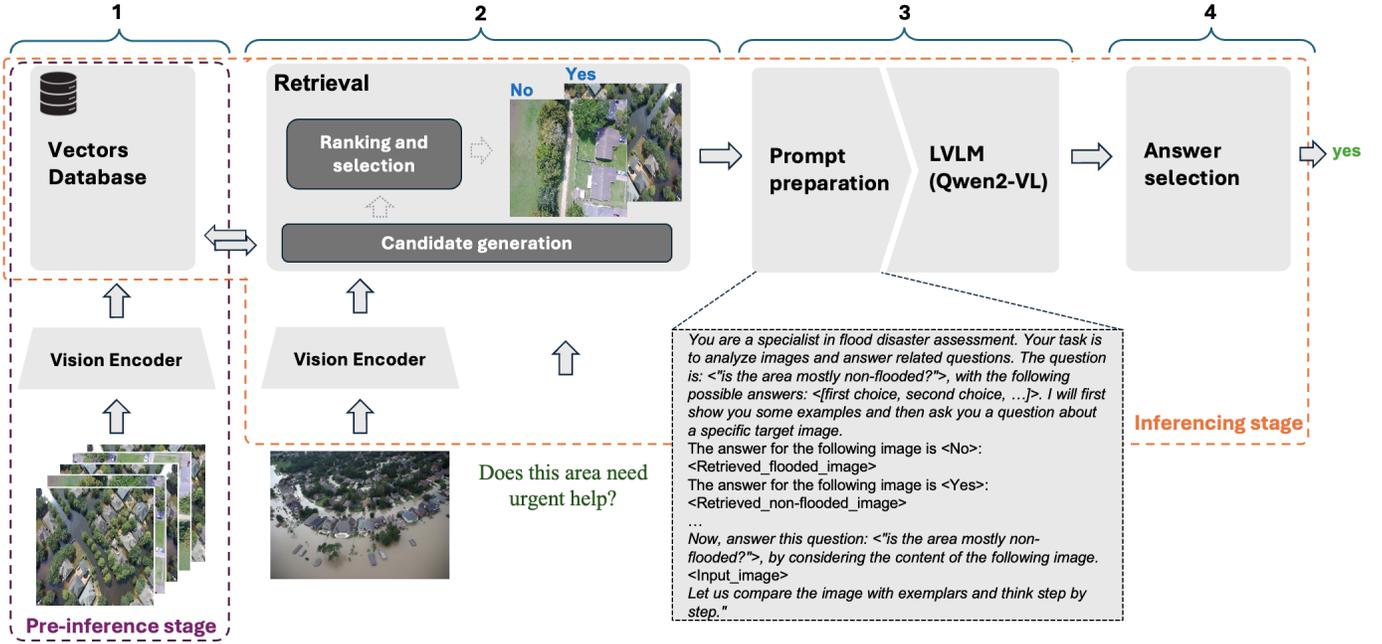

Fig. 4. Overview of the proposed framework (ThiFAN-VQA) for disaster-related VQA. The framework architecture consists of four stages: (1) training set encoding, where training image embeddings are extracted and stored in a vector database; (2) exemplar retrieval, which selects similar image–answer pairs based on the question type, candidate answers, and visual features; (3) prompt generation and answer prediction, where retrieved exemplars are inserted into a CoT-style prompt for answer prediction using a MLLM; and (4) answer verification and selection, which ensures that the output aligns with predefined answer candidates for multiple-choice questions.

After ranking, exemplar selection is performed according to the type of question:

- **Multiple-choice questions:** For each candidate answer, the most visually similar exemplar is retrieved. This ensures that the final set of exemplars represents all possible answers.
- **Counting questions (simple or complex):** The 2-top visually similar samples are selected directly from the filtered candidate pool, without considering specific answer categories or numbers.

For example, in binary (yes/no) condition-recognition tasks, two exemplars are retrieved—one labeled "yes" and one labeled "no"—each corresponding to the image most visually similar to the query image within its respective class. The resulting image–answer pairs serve as in-context demonstrations for inference.

*3) Prompt generation and reasoning generation:* At this stage, the retrieved demonstration exemplars are integrated into a prompt template, which is structured based on the selected retrieval strategy. For instance, the prompt for a question from the entire_image_condition category—a binary (yes/no) question—is formatted as follows:

*"You are a specialist in flood disaster assessment. Your task is to analyze images and answer related questions. The question is: <"is the area mostly non-flooded?">, with the following possible answers: <[Yes, No]>. I will first show you some examples, and then ask you a question about a specific target image.*
**The answer for the following image is <No>:*
<Retrieved_flooded_image>
**The answer for the following image is <Yes>:*
<Retrieved_non-flooded_image>
*Now, answer this question: <"is the area mostly non-flooded?">, by considering the content of the following image.*
<Input_image>
*Let us compare the input image with exemplars and provide me with the reasoning step by step."*

For counting-type questions, the prompt differs from the multiple-choice format by omitting the list of candidate answers. It presents only a set of 2 exemplar image–answer pairs to help the model infer the underlying reasoning pattern.

The prompts are designed to elicit CoT reasoning [44], [57], encouraging the model to generate intermediate reasoning steps before producing a final answer. Specifically, we adopt a zero-shot CoT prompting by appending the phrase *"Let us compare the input image with exemplars and provide me with the reasoning step by step."* which guides the LVLM to explicitly articulate its comparison and reasoning process.

*4) Answer selection:* With the exception of the simple-counting and complex-counting categories, all other question types are framed as multiple-choice tasks. Therefore, during the prompt generation and answer selection stage (Section IV-A3), the MLLM-generated answer must be mapped to one of the candidate answers provided in the flexible list embedded in the prompt. Unlike prior work such as [14], which relies on post-hoc similarity matching, we explicitly query an LLM and instruct it to select one of the acceptable answer choices based solely on the intermediate reasoning, ignoring the model's initial final prediction. This approach helps reduce model hallucination and improves the overall reliability of the framework [58]. The following prompt template is used for this purpose:

*"The answer that has been generated for the question <"Question" > is: <"Answer" >. Based on the reasoning, select one of the following choices without adding any additional information: <[Candidate_answers] >."*

For counting-type questions, the model is directly asked to produce a free-form numerical prediction without a predefined list of numbers. Fig. 5 illustrates an example of the model's prediction process. First, the MLLM is provided with two exemplars, the question, and the corresponding image, and is asked to generate the reasoning. The sequence of intermediate reasoning outputs shows how the MLLM initially processes the visual input and attempts to count the buildings, first producing an incorrect estimate of 8, even though the intermediate reasoning itself is correct. Subsequent step reveals the internal reasoning and demonstrates how the reasoning-based answer selection module evaluates and corrects this initial hallucination, ultimately producing the correct answer.

## V. RESULTS

We begin by evaluating ThiFAN-VQA on the FloodNet dataset [4] and comparing its performance with the zero-shot ZeShot-VQA baseline [14]. After establishing this comparison on FloodNet, we extend the analysis to the RescueNet-VQA dataset to further assess the relative performance of the two models across additional question types. Finally, we compare ThiFAN-VQA with supervised approaches, including SAM-VQA [9] and SAN [20], to examine how our ICL-based method performs against fully supervised models.

As shown in Table I, both ZeShot-VQA and our proposed method use the same input modalities—image, question, and answer triplets (I-Q-A)—and neither requires segmentation masks. Both models do not need to be trained on a new dataset, and they have a flexible answer space because they select a prediction from a list of candidate answers provided at inference time. ThiFAN-VQA leverages in-context learning (ICL) with 798 unique images as exemplars, whereas ZeShot-VQA operates in a purely zero-shot setting without any training images. Overall, our method achieves an accuracy of 77%, substantially outperforming the zero-shot ZeShot-VQA baseline (62%) across all question categories. To ensure a fair comparison under identical conditions, we also conducted a zero-shot evaluation of our proposed method. The results, presented in the first and second columns of the Table I, show that our model outperforms ZeShot-VQA across all multiple-choice question types—except for the density estimation task—even without the use of exemplars.

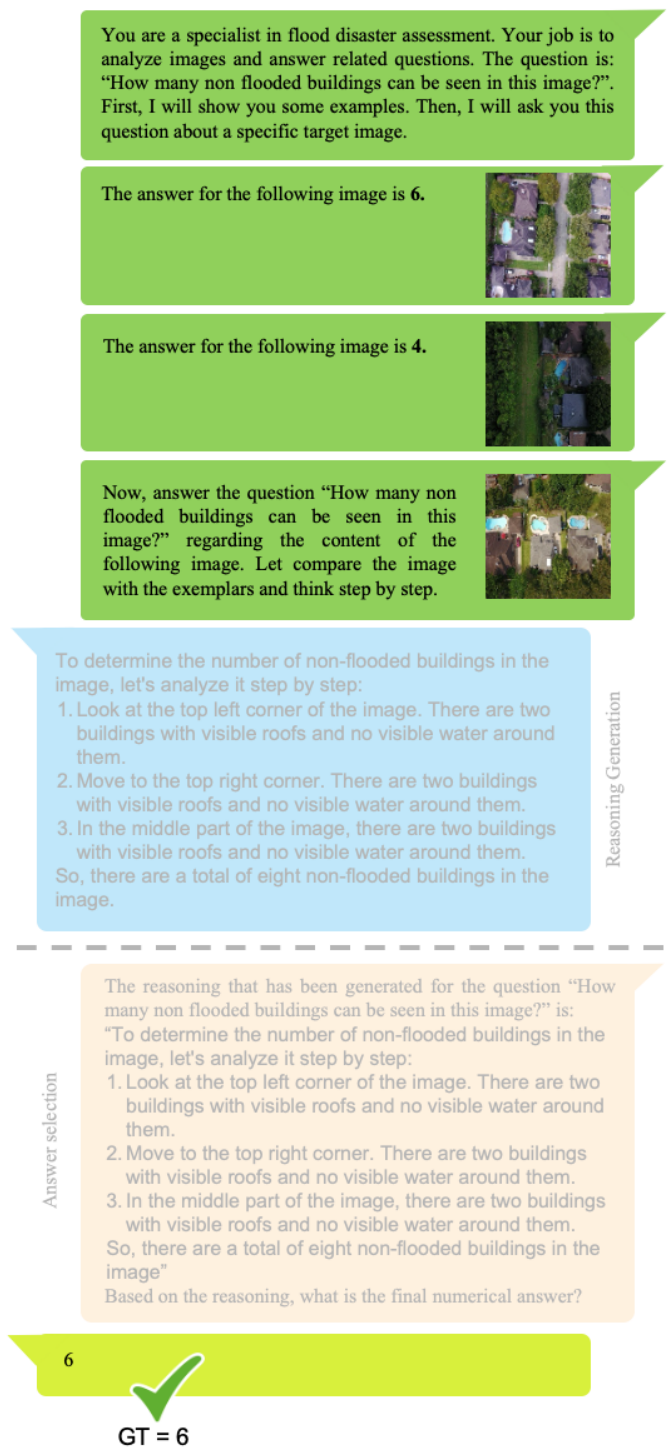

Fig. 5. Example illustrating the outputs of ThiFAN-VQA, including intermediate reasoning and the final prediction for a counting-type question using CoT reasoning. The prompt includes two in-context examples that guide the LVLM to reason step-by-step. In this particular example, although the model initially hallucinates a count of 8, the reasoning-guided answer selection module corrects this error and produces the accurate final answer of 6.

TABLE I
COMPARISON BETWEEN ZESHOT-VQA AND THE PROPOSED METHOD ACROSS VARIOUS VQA TASKS FROM THE FLOODNET DATASET. BOTH METHODS DO NOT REQUIRE RETRAINING, FINE-TUNING, SEGMENTATION MASKS, AND ARE NOT RESTRICTED TO A CLOSED ANSWER SPACE. WHILE ZESHOT-VQA OPERATES IN A FULLY ZERO-SHOT SETTING, THE PROPOSED METHOD LEVERAGES ICL WITH 798 EXEMPLAR IMAGES. THE PROPOSED METHOD CONSISTENTLY OUTPERFORMS ZESHOT-VQA ACROSS ALL QUESTION TYPES, PARTICULARLY IN MULTIPLE-CHOICE CATEGORIES, WHILE RETAINING THE GENERALIZATION CAPABILITIES OF ZERO-SHOT APPROACHES.

|  | ZeShot-VQA [14] | ThiFAN-VQA (zero-shot) | ThiFAN-VQA (Full proposed model) |
| --- | --- | --- | --- |
| Data | I-Q-A | I-Q-A | I-Q-A |
| Method | Zero-shot | Zero-shot | ICL |
| Training free | ✓ | ✓ | ✓ |
| Flexible answer space | ✓ | ✓ | ✓ |
| # unique images from the support set | 0 | 0 | 798 |
| Question Type |  |  |  |
| Building Condition Recognition | 0.78 | 0.88 | **0.95** |
| Density Estimation | 0.38 | 0.36 | **0.62** |
| Entire Image Condition Recognition | 0.59 | 0.86 | **0.97** |
| Risk Assessment | 0.52 | 0.88 | **0.93** |
| Road Condition Recognition | 0.41 | 0.80 | **0.95** |
| Complex Counting | 0.22 | 0.18 | **0.27** |
| Simple Counting | 0.27 | 0.25 | **0.30** |
| Overall Accuracy | 0.62 | 0.66 | **0.77** |

TABLE II
COMPARISON OF ACCURACY BETWEEN ZESHOT-VQA AND THE PROPOSED METHOD ACROSS VARIOUS VQA TASKS FROM THE RESCUNET-VQA DATASET.

| Question Type | ZeShot-VQA | ThiFAN-VQA |
| --- | --- | --- |
| Building Condition Recognition | 0.60 | **0.68** |
| Density Estimation | 0.72 | **0.75** |
| Area based | 0.33 | **0.38** |
| Risk Assessment | **0.81** | 0.81 |
| Road Condition Recognition | 0.49 | **0.72** |
| Complex Counting | 0.16 | **0.30** |
| Simple Counting | 0.19 | **0.31** |
| Level of damage | 0.41 | **0.64** |

Beyond the results on the FloodNet dataset reported in Table I, Table II provides a performance comparison of ZeShot-VQA and ThiFAN-VQA on the RescueNet-VQA dataset across all question types. The results indicate that the proposed model consistently outperforms ZeShot-VQA for every VQA task. However, both simple counting and complex counting remain challenging due to the inherent difficulty of numerical reasoning, which continues to be a general limitation across VQA frameworks.

The comparison of ThiFAN-VQA with the supervised models SAM-VQA and SAN on the FloodNet dataset is reported in Table III. SAM-VQA and SAN are fully supervised models trained on the complete set of I–Q–A triplets, with SAM-VQA additionally leveraging segmentation masks as auxiliary supervision for flood disaster damage assessment. Overall, our proposed model (ThiFAN-VQA) performs competitively with SAN, which is trained with the same data modalities, while using significantly fewer unique exemplar images than the amount of training data required by SAN. Moreover, both SAM-VQA and SAN are restricted to the answer space defined by their training sets, whereas our proposed method supports a flexible answer space. Furthermore, our method outperforms both supervised models in the Building Condition Recognition task. These results highlight the effectiveness of our ICL-based framework in narrowing the performance gap between zero-shot generalization and supervised learning for disaster-related VQA tasks.

*A. Results analysis*

Considering practical constraints such as dataset preparation time and model limitations, our method achieves competitive performance across both counting and multiple-choice question categories—including Building Condition, Entire Condition, Risk Assessment, and Road Condition—when compared to the supervised SAN model trained on the same data modalities. These results underscore the effectiveness of MLLMs when paired with carefully selected in-context exemplars, CoT prompting, and an the reasoning-based answer selection module. Notably, our approach attains the highest accuracy in the Building Condition category (95%), outperforming all baselines (see Table III).

However, the model underperforms on Density Estimation, Simple Counting, and Complex Counting tasks compared to SAM-VQA supervised baseline. We attribute this performance gap to two main factors. First, SAM-VQA benefits from the use of semantic segmentation masks during training, which provide strong supervision for object localization and scene understanding—capabilities that are particularly valuable for density estimation and counting tasks such as estimating the number of flooded buildings. However, this reliance on segmentation annotations limits the practicality of SAM-VQA in real-world urgent scenarios. Generating accurate masks is a time-consuming and labor-intensive process, making it impractical for rapid deployment during emergencies, especially in the immediate aftermath of a new disaster. Second, counting questions present an inherently open-ended prediction challenge, as they lack the constrained answer space found in multiple-choice formats. This makes them more difficult for the proposed model, which generates free-form answers, compared to supervised classification-based methods. Supervised models use closed-vocabulary decoders, which limit the output space to a fixed set of predefined answers.

TABLE III
COMPARISON OF VQA ACCURACY ON THE FLOODNET DATASET BETWEEN SUPERVISED METHODS (SAM-VQA, SAN) AND THE PROPOSED ICL-BASED APPROACH. SAM-VQA AND SAN ARE SUPERVISED MODELS, WITH SAM-VQA ADDITIONALLY LEVERAGING SEGMENTATION MASKS AS AUXILIARY SUPERVISION. IN CONTRAST, THE PROPOSED METHOD OPERATES WITHOUT ANY RETRAINING OR FINE-TUNING, DOES NOT REQUIRE SEGMENTATION MASKS, AND IS NOT LIMITED TO A CLOSED ANSWER SPACE.

|  | SAM-VQA [9] | SAN [9] | ThiFAN-VQA |
|---|---|---|---|
| Data | I-Q-A + segmentation mask | I-Q-A | I-Q-A |
| Method | Supervised | Supervised | ICL |
| training-free | ✗ | ✗ | ✓ |
| Flexible answer space | ✗ | ✗ | ✓ |
| # images in training set/support set | 1364 | 1364 | **798** |
| Question Type |  |  |  |
| Building Condition Recognition | 0.94 | 0.93 | **0.95** |
| Density Estimation | **0.74** | 0.68 | 0.62 |
| Entire Image Condition Recognition | **0.98** | **0.98** | 0.97 |
| Risk Assessment | **0.97** | **0.97** | 0.93 |
| Road Condition Recognition | **0.98** | **0.98** | 0.95 |
| Complex Counting | **0.35** | 0.28 | 0.27 |
| Simple Counting | **0.35** | 0.30 | 0.30 |
| Overall Accuracy | **0.81** | 0.80 | 0.77 |

TABLE IV
IMPACT OF EXEMPLAR COUNT PER ANSWER CHOICE ON VQA ACCURACY IN THE FLOODNET DATASET. INCREASING THE NUMBER OF I-Q-A EXEMPLAR CANDIDATES PER CHOICE LEADS TO CONSISTENT PERFORMANCE GAINS ACROSS ALL QUESTION TYPES.

| Number of candidate I-Q-A / choice | 0 | 3 | 5 | 7 | totall |
|---|---|---|---|---|---|
| Number of exemplars / choice | 0 | 1 | 1 | 1 | 1 |
| # unique images from support set | 0 | 149 | 203 | 248 | 798 |
| Question Type |  |  |  |  |  |
| Building Condition Recognition | 0.86 | 0.88 | 0.93 | 0.93 | **0.95** |
| Density Estimation | 0.35 | 0.36 | 0.36 | 0.40 | **0.62** |
| Entire Image Condition Recognition | 0.82 | 0.86 | 0.90 | 0.91 | **0.97** |
| Risk Assessment | 0.74 | 0.88 | 0.93 | **0.94** | 0.93 |
| Road Condition Recognition | 0.67 | 0.80 | 0.80 | 0.85 | **0.95** |
| Complex Counting | 0.13 | 0.18 | 0.20 | 0.24 | **0.27** |
| Simple Counting | 0.18 | 0.25 | 0.25 | 0.28 | **0.30** |
| Overall Accuracy | 0.59 | 0.66 | 0.69 | 0.71 | **0.77** |

This restriction simplifies the prediction task by narrowing the model's decision space, often resulting in higher accuracy for structured tasks such as counting [59].

Figure 6 presents qualitative results from the FloodNet dataset, demonstrating the model's reasoning performance across various question categories. In simple and complex counting tasks (e.g., "How many buildings can be recognized as non-flooded?"), the model generally identifies the number of structures with high accuracy, and in many instances, its predictions closely match the ground truth. However, even minor numerical deviations are counted as incorrect, which negatively impacts the evaluation metrics. For density estimation and entire-image condition recognition questions, the model effectively captures the overall flood severity and scene status, such as distinguishing flooded from non-flooded regions. In risk assessment questions (e.g., "Do the rescuers need to take immediate actions?"), the framework provides reliable and contextually appropriate responses consistent with the ground truth. Likewise, in road condition recognition tasks, it correctly identifies flooded roads and captures essential situational cues. Overall, these examples illustrate that the proposed framework can generalize well across diverse reasoning types, producing accurate and interpretable answers even without task-specific training.

*B. Ablation study*

To evaluate the effect of dataset size on the proposed ICL framework, we conduct an ablation study by varying the number of available I-Q-A triplets per answer choice, while keeping the number of in-context exemplars fixed at one per choice. As shown in Table IV, increasing the number of available triplets improves performance in almost all types of questions.

Even with as few as three triplets per choice, the model shows noticeable gains over the zero-shot baseline. For instance, accuracy on Entire Image Condition Recognition improves from 43% to 86%, and Risk Assessment rises from 68% to 88%. Further increasing the image count to five and seven per choice continues to boost accuracy, though with diminishing returns for some tasks. The performance plateaus slightly for Risk Assessment and Building Condition Recognition, while more substantial improvements are observed in Density Estimation and Road Condition Recognition, where visual variation may require more training diversity.

Notably, Simple Counting and Complex Counting remain the most challenging tasks [25], showing only modest gains even with more data. This aligns with earlier observations that open-ended counting tasks benefit less from additional exemplars compared to multiple-choice categories. Overall, the model achieves an accuracy of 77% when the model can see all available I-Q-A triplet candidates, highlighting the importance of exemplar diversity in the ICL setup.

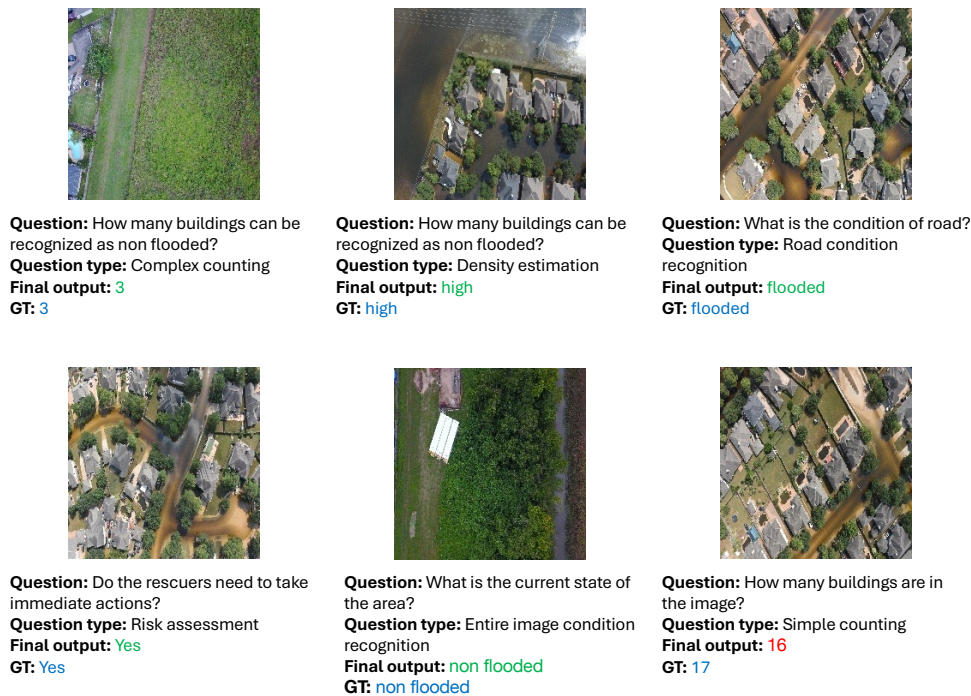

Fig. 6. Qualitative results of the proposed framework on the FloodNet dataset. Each example shows the input UAV image and corresponding question, along with the model's predicted answer, ground truth (GT), and question type.

TABLE V
THE EFFECT OF THE SELECTION STAGE AND THE CoT ON THE MODEL PERFORMANCE

| Method | Descriptioin | Accuracy |
|---|---|---|
| Proposed | w/o Answer selection and CoT | 0.59 |
| Proposed | w/o CoT | 0.74 |
| Proposed |  | **0.77** |

To evaluate the contribution of key components in our framework, we perform ablations by removing (1) the answer selection stage and (2) the CoT prompting. As shown in Table V, both components play a critical role in the model's overall performance.

Removing the answer selection mechanism causes a substantial drop in accuracy—from 77% to 59%—highlighting its importance in aligning the LVLM's free-form output with predefined candidate answers in multiple-choice settings. Similarly, disabling CoT prompting results in a decrease in accuracy to 74%, indicating that encouraging the model to reason step-by-step contributes positively to its prediction quality.

## VI. CONCLUSION

In this work, we presented ThiFAN-VQA, a reasoning-based VQA framework designed for Visual Question Answering (VQA) in post-disaster damage assessment. The proposed approach effectively bridges the gap between supervised and zero-shot methods by integrating In-Context Learning (ICL) with a customized retrieval method and comparison-based chain-of-thought (CoT) prompting within a unified framework. To mitigate hallucination and reasoning–answer inconsistency common in multi-modal large language models (MLLMs), ThiFAN-VQA introduces a two-stage pipeline that separates reasoning generation from answer selection, ensuring that the final predictions remain coherent with the inferred reasoning process. Comprehensive experiments on post-disaster aerial-based FloodNet and RescueNet-VQA datasets demonstrate that our framework achieves competitive or superior performance compared to both zero-shot and fully supervised baselines, without requiring retraining, fine-tuning, or auxiliary inputs such as segmentation masks.

In general, ThiFAN-VQA provides a rapid deployable, flexible, and interpretable solution for automated post-disaster scene understanding, enabling decision-makers to extract actionable insights from aerial imagery more efficiently. Future research will explore extending the framework to multi-temporal and multimodal settings, integrating temporal reasoning for change detection, and leveraging self-adaptive retrieval mechanisms to further enhance scalability and autonomy in real-world disaster response systems.

## ACKNOWLEDGMENT

This work is partially supported by the National Science Foundation (grant #2423211).